\pdfoutput=1

\documentclass[11pt]{article}

\usepackage{EMNLP2022}

\usepackage{times}
\usepackage{latexsym}

\usepackage[T1]{fontenc}

\usepackage[utf8]{inputenc}

\usepackage{microtype}

\usepackage{inconsolata}

\usepackage{graphicx}
\usepackage{bm}
\usepackage{amssymb}
\usepackage{amsmath}
\usepackage{multirow}
\usepackage[normalem]{ulem}
\useunder{\uline}{\ul}{}

\usepackage{enumitem}
\usepackage{pifont}

\newcommand{\ie}{\textit{i}.\textit{e}.}
\newcommand{\eg}{\textit{e}.\textit{g}.}

\newcommand{\etc}{\textit{etc}.}

\title{Rethinking Multi-Modal Alignment in Multi-Choice VideoQA from Feature and Sample Perspectives}

\author{Shaoning Xiao$^\heartsuit$, \! Long Chen$^\spadesuit$\thanks{~~Long Chen is the corresponding author.}, \! Kaifeng Gao$^\heartsuit$, \! Zhao Wang$^\heartsuit$, \!
Yi Yang$^\heartsuit$, \\ \textbf{Zhimeng Zhang$^\heartsuit$, \! and Jun Xiao$^\heartsuit$} \\
$^\heartsuit$Zhejiang University \;
$^\spadesuit$Columbia University \\
\texttt{\{shaoningx,kite\_phone,zhao\_wang,yangyics,zhimeng,junx\}@zju.edu.cn}, \\
\texttt{zjuchenlong@gmail.com}}

\begin{document}
\maketitle
\begin{abstract}
Reasoning about causal and temporal event relations in videos is a new destination of Video Question Answering (VideoQA).
The major stumbling block to achieve this purpose is the semantic gap between 
language and video since they are at different levels of abstraction. 
Existing efforts mainly focus on designing sophisticated architectures while utilizing frame- or object-level visual representations. 
In this paper, we reconsider the multi-modal alignment in VideoQA from feature and sample perspectives to achieve better performance.
From the view of feature,
we break down the video into trajectories and first leverage trajectory feature in VideoQA to enhance the alignment between two modalities. 
Moreover, we adopt a heterogeneous graph architecture and design a hierarchical framework to align both trajectory-level and frame-level visual feature with language feature.
In addition, we found that VideoQA models are largely dependent on language priors and always neglect visual-language interactions.
Thus, two effective yet portable training augmentation strategies are designed to strengthen the cross-modal correspondence ability of our model from the view of sample.
Extensive results show that our method outperforms all state-of-the-art models 
on the challenging NExT-QA benchmark.

\end{abstract}

\section{Introduction}
Given a video and a question about its content, Video Question Answering (VideoQA) aims to answer the question through multi-modal reasoning. Since it can benefit numerous multi-modal applications such as video retrieval~\cite{DBLP:conf/sigir/0048SJNZZG21,DBLP:conf/aaai/XiaoCZJSYX21,xiao2021natural} and interaction with robot vision, VideoQA has received increasing attention in recent years.   
Compared to Image Question Answering (ImageQA)~\cite{DBLP:conf/iccv/AntolALMBZP15}, VideoQA requires more complex reasoning. 
A naive extension of ImageQA methods~\cite{chen2020counterfactual,chen2022rethinking} may not apply to the problem of VideoQA due to the extra information such as temporal object interactions.

\begin{figure}
    \centering
    \includegraphics[width=0.48\textwidth]{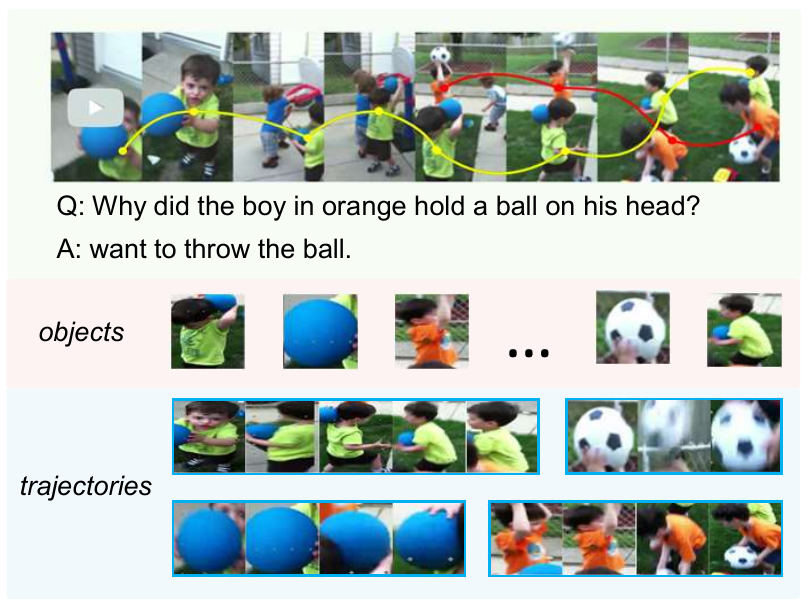}
    \caption{An illustration of VideoQA. Existing VideoQA models utilize objects as finer clues. We further track the objects with same classes across time and first leverage video trajectory for better alignment.}
    \label{fig1}
\end{figure}

As computing systems are more frequently intervening to improve people’s work and daily lives,
it is critical for machines to correctly comprehend the 
complex relationships between events such as 
causal and temporal relations.
However, conventional VideoQA benchmarks~\cite{DBLP:conf/mm/XuZX0Z0Z17,DBLP:conf/aaai/YuXYYZZT19} are constituted by
descriptive questions, \eg, asking the number, color or single action of the video elements. 
Recently, a benchmark called NExT-QA~\cite{DBLP:conf/cvpr/XiaoSYC21} was elaborately designed to challenge VideoQA models to reason about causal and temporal actions and understand the rich object interactions in daily activities.
Instead of a fixed answer set, NExT-QA requires the models to pick the correct answer out of five candidates. Since this multi-choice setting allows concatenating question-answer pair to form a holistic language query and aligning it with videos, it can be easily generalized to any other multi-modal tasks besides question answering.

Multi-modal alignment, namely finding the correspondences between two modalities, is the foundation of multi-modal tasks.
One of the major challenges to align two modalities is to fill in the semantic gap between language and video due to different levels of abstraction.
Intuitively, the key to better alignment is to identify the direct relations between sub-elements (\eg, words, frames, \etc) of instances from two different modalities.
As a highly abstract vehicle of expression, almost every word in a natural language sentence can express a complete meaning, \eg, an entity, an action or a time.
In contrast, a video also contains a number of visual concepts but they are naturally indivisible unlike the word. 
Primitive VideoQA models~\cite{DBLP:conf/aaai/LiSGLH0G19,DBLP:conf/cvpr/JangSYKK17} utilize frame-level appearance feature as video representation and align it with language feature. These methods simply regard the video as a series of frames. 
To better capture dynamic change in videos, others~\cite{DBLP:conf/aaai/GaoZSLLMS19,DBLP:conf/cvpr/FanZZW0H19,DBLP:conf/aaai/JiangH20} incorporate frame-level motion feature together with appearance feature. 
Recently, object feature as more granular information has been leveraged to strengthen the semantic correspondence ability~\cite{DBLP:conf/aaai/HuangCZDTG20,DBLP:conf/cvpr/LeLV020,DBLP:HQ-GAU}. 
Although bringing in object information results in a better ability to reason about complex interactions in videos, there are still two unsolved problems: (1) Static object information is hard to model temporal-related relations. (2) Same object may take different actions during time and different objects with the same label can cause a mismatch. 
As shown in Figure~\ref{fig1}, there are two boys and two balls in the video and both boys hold a ball. In this case, VideoQA models that only use object-level information may end up with misalignment, which leads to a wrong answer.
Therefore, tracking objects across time is of vital importance. 

In this paper, we explore the multi-modal alignment in VideoQA from feature and sample perspectives to better reason over videos’ causal/temporal actions. 
From feature perspective, we decouple trajectories as salient entities from video and first leverage video trajectory feature in VideoQA. In order to model the rich interactions between trajectories, we propose a trajectory encoder using multi-head self-attention with temporal and semantic embeddings. 
Video trajectories are the essential ingredient for video relation detection task~\cite{DBLP:conf/mm/QianZLXP019,DBLP:conf/mm/XieR020}, which require tracking the same object from different frames along the temporal axis. 
Specifically, we first apply a pre-trained object detector to obtain bounding boxes. Then, an association algorithm named improved sequence NMS~\cite{DBLP:conf/mm/XieR020} is applied to obtain video trajectories that contain spatial-temporal information of the visual elements. 
We further align trajectory-level and frame-level feature with language feature by a cycle-attention module and adopt a heterogeneous graph architecture for implicit relation reasoning.

In addition, from sample perspective, we design two training strategies in order to enhance multi-modal alignment in feature space. To be specific, we first increase negative candidate answers when computing the matching score. This strategy forces the model to focus on the discriminative regions within a question-answer pair.
We then add negative question-answer pairs that are attached to other videos.
By doing so, the video and its affiliated language are drawn closer in feature space and the mismatched pairs are pulled away.
Moreover, we found that these strategies can also solve the problem that VideoQA models are largely dependent on language priors and neglect visual-language interactions.
Together with the proposed model, our method achieved state-of-the-art performance.

In summary, the main contributions of our work are listed as follows:
(1) We first leverage video trajectory features in VideoQA to capture richer causal and temporal relations in the video. 
(2) We design two training strategies to strengthen the cross-modal correspondence ability of our model and further boost the performance. 
(3) We conducted extensive experiments on NExT-QA and the results demonstrate the effectiveness of our model.

\begin{figure*}
    \centering
    \includegraphics[width=1.0\textwidth]{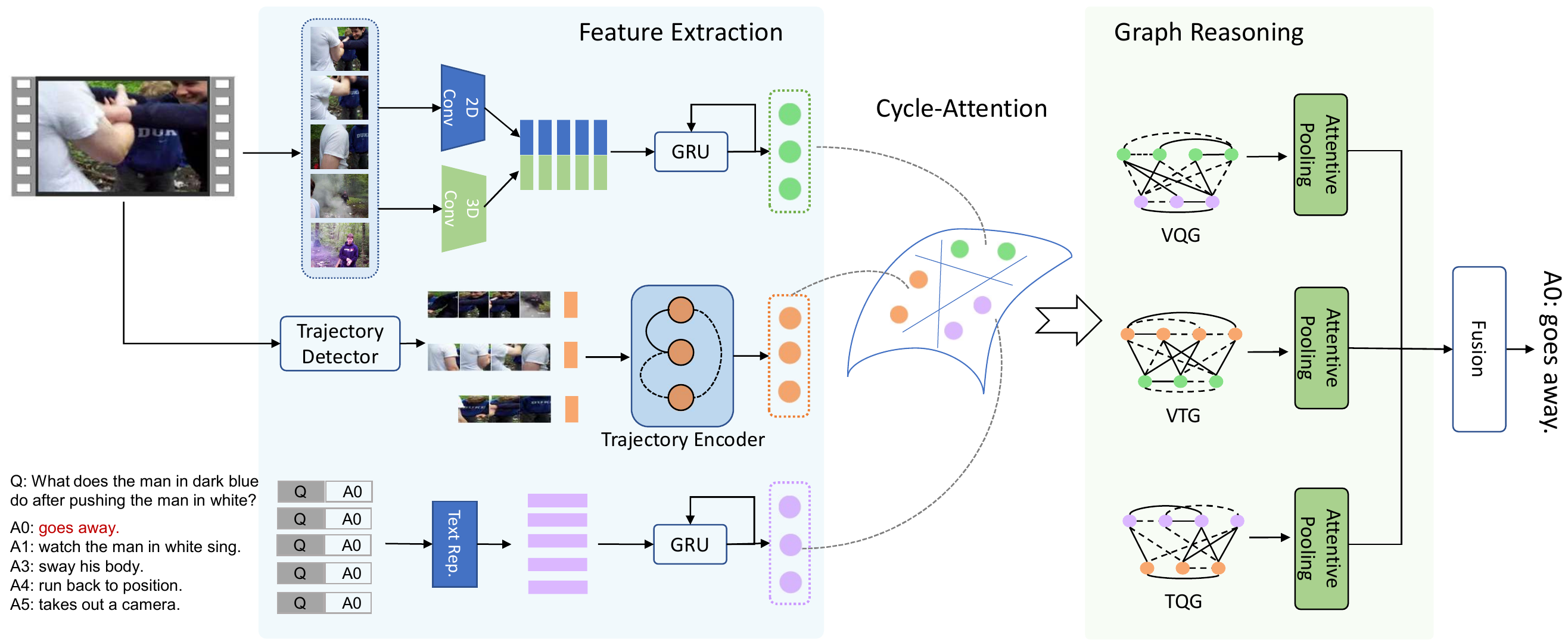}
    \caption{The overview of our model architecture for VideoQA. Firstly, the frame-level features, trajectory-level features and text representations are extracted. Then, the visual and language features are aligned in pairs by a cycle-attention module. At last, heterogeneous graphs are constructed and applied for reasoning.}
    \label{fig2}
\end{figure*}

\section{RELATED WORK}
\subsection{Video Question Answering}
We roughly summarise three kinds of VideoQA methods according to their utilized techniques, namely attention-based, memory-based, and graph-based models. Attention mechanism~\cite{DBLP:conf/cvpr/JangSYKK17,ye2017video,DBLP:conf/aaai/GaoZSLLMS19,DBLP:conf/aaai/LiSGLH0G19,DBLP:conf/aaai/JiangCLZG20} is widely used in VideoQA.
\citet{DBLP:conf/cvpr/JangSYKK17} propose a dual-layer LSTM with spatial
and temporal attention.
\citet{DBLP:conf/aaai/LiSGLH0G19} use the self-attention mechanism to encode each modality and utilize co-attention mechanism for alignment.
\citet{DBLP:conf/aaai/JiangCLZG20} divide the semantic features generated from question into  the spatial part and the temporal part which guide the spatial and temporal attention of video, respectively.
Memory based approaches~\cite{DBLP:conf/mm/XuZX0Z0Z17,DBLP:conf/cvpr/GaoGCN18,DBLP:conf/cvpr/FanZZW0H19} encode the input sources multiple cycles and use attention mechanism allowing the model to focus on different contents in each cycle.
\citet{DBLP:conf/mm/XuZX0Z0Z17} gradually refine the attention over the appearance and motion features of the video using the question as guidance. 
\citet{DBLP:conf/cvpr/GaoGCN18} propose a co-memory attention module to extract useful cues from both appearance and motion memories to generate attention for motion and appearance separately.
\citet{DBLP:conf/cvpr/FanZZW0H19} propose a read-write memory network that jointly encode the movie appearance and caption content.

Graph based models~\cite{DBLP:conf/iccv/HuRDS19,DBLP:conf/aaai/HuangCZDTG20,DBLP:conf/cvpr/ParkLS21,DBLP:conf/aaai/JiangH20,DBLP:HQ-GAU} advance the field by exploiting the ability of relation reasoning. 
\citet{DBLP:conf/aaai/JiangH20} propose a heterogeneous graph alignment network to align and interact the inter- and intra-modality. 
\citet{DBLP:conf/cvpr/ParkLS21} perform relation reasoning between appearance and motion information of the video with compositional semantics of the question. 
Although these methods achieve impressive performance using the graph structure, they only utilize frame-level video feature for alignment and thus suffer from a lack of fine-grained interaction. Some other methods leverage  object information to enhance the fine-grained alignment.
\citet{DBLP:conf/iccv/HuRDS19} propose a graph network where each node represents an object,  
and conduct iterative message passing conditioned on the textual input.
\citet{DBLP:conf/aaai/HuangCZDTG20} propose to represent video as a location aware graph 
and conduct graph convolution.
\citet{DBLP:HQ-GAU} model video as a conditional hierarchical graph to align the video facts and textual cues on different levels. 
Instead of designing complicated models, we solve VideoQA task by leveraging a new trajectory feature and further boost performance from sample perspective. 
\subsection{Video Trajectory Detection}
Video trajectory detection, as an essential component of video relation detection task~\cite{DBLP:conf/mm/QianZLXP019,DBLP:conf/mm/XieR020,gao2022classification,gao2021video}, has attracted more and more attention.
Detection of objects in static image has gained a great improvement in last few years. 
However, video trajectory detection is still a tough problem since it needs to tracking
same object in different frames of a video clip. 
A popular scheme is tracking-by-detection, namely applying detection algorithm to each video frame and the detections are associated across frames to form trajectories.
Seq-NMS~\cite{DBLP:journals/corr/HanKPRBSLYH16} takes detections from a state-of-the-art object detection method and associates over time by finding the highest scoring path. 
Improved Seq-NMS~\cite{DBLP:conf/mm/XieR020} improves seq-NMS by introducing a new linking mechanism to solve the missing connection problem caused by violent object movement. 
In this paper, we detect static objects using a pre-trained detector and utilize improved seq-NMS as trajectory tracking method to generate trajectories. 

\section{Approach}

Formally, suppose we have a video $V={\{v_t\}}^T_{t=1}$ which contains $T$ frames and $v_t$ denotes the $t$-th frame. Meanwhile, we have a natural language question $Q={\{w_l\}}^L_{l=1}$,
where $w_l$ denotes the $l$-th word in the sentence and $L$ represents the question length. 
VideoQA aims to predict the correct answer $A^{p}$ to the question according to the relevant video content. In the multi-choice setting, the goal is to choose the correct answer $A^{p}$ from $n$ candidate answer set $S_A = \{A_1, A_2,...,A_n\}$.

In this section, we sequentially introduce each component of our proposed model.
The video and language encoding procedures are presented in Section~\ref{sec3.1}.
The alignment and reasoning modules are introduced in Section~\ref{sec3.2}. The answer predictor is introduced in Section~\ref{sec3.3}.
In Section~\ref{sec3.4}, we introduce two sample augmentation strategies.

\subsection{Feature Encoding}
\label{sec3.1}

\subsubsection{Video Representations} 
We utilize both frame-level and trajectory-level video features for video representation since they naturally share complementary information.

\noindent\textbf{Frame-level Features.} Following previous works, we uniformly sample a fixed number {\small$N$} of clips for each video. 
We use a 2D ConvNet to extract video appearance feature {\small$ \bm{F_a}={\{f^a_i\}}^N_{i=1}$}, where {\small$f^a_i\in\mathbb{R}^{d_a}$} and use a 3D ConvNet to extract video motion feature {\small$\bm{F_m}={\{f^m_i\}}^N_{i=1}$}, where {\small$f^m_i\in\mathbb{R}^{d_m}$}. 

Then, we apply a concatenation operation for the appearance and the motion feature with a fully-connected layer to obtain frame-level video feature,
{\small$\bm{F}_v=ReLU(FC([\bm{F}_a, \bm{F}_m]))$},
where {\small$\bm{F}_v\in\mathbb{R}^{N \times d}$} and {\small$[\cdot]$} represent the concatenation operation along the feature dimension.
Due to the temporal property of videos, we adopt a Gated Recurrent Units~\cite{DBLP:conf/emnlp/ChoMGBBSB14} to process the frame-level video feature,
{\small$\bm{V}=GRU(\bm{F}_v)$},
where {\small$\bm{V}\in\mathbb{R}^{N \times d}$} is the contextualized frame-level video features.

\noindent\textbf{Trajectory-level Features.}
As mentioned before, we argue that the video and question are at different abstract levels due to their sub-components, \ie, words and frames contain inconsistent semantic information. Thus, we utilize video trajectory feature to supplement the frame-level feature in order to enhance the feature alignment with language.

\begin{figure}
    \centering
    \includegraphics[width=0.45\textwidth]{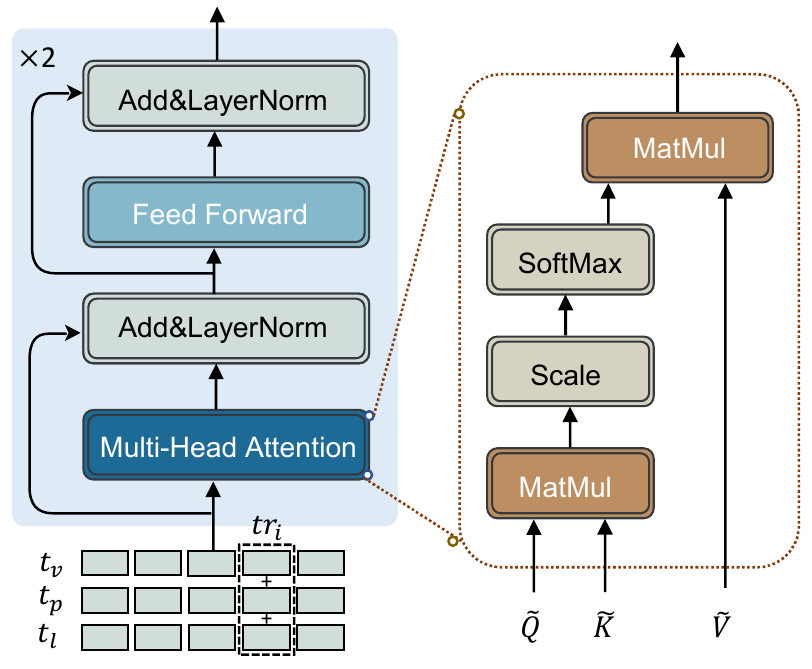}
    \caption{The architecture of our trajectory encoder. Right part is an illustration of dot-product attention.}
    \label{fig3}
\end{figure}

We take the tracking-by-detection strategy to generate video trajectories. 
We first sample the video and detect the objects from all the frames. 
To generate trajectories, we use improved seq-NMS~\cite{DBLP:conf/mm/XieR020} to associate bounding boxes along the time that belong to same object based on object detection results. 
Specifically, this algorithm links the bounding boxes that likely belong to the same object from consecutive frames to build a graph and it applies dynamic programming to repeatedly pick the path with the highest score. 
Then, we obtain a series of trajectories each of which contains a set of boxes, a predicted label and the start-end time points. 
For each trajectory, we apply average pooling to the associated objects features and normalize the start-end time with respect to the video length. 
To take advantage of semantic information, we project the trajectory label to semantic space using GloVe embeddings~\cite{DBLP:conf/emnlp/PenningtonSM14}.
Thus, we obtain the visual feature {\small$t_v$}, semantic feature {\small$t_l$} and temporal position embedding {\small$t_p$} for each trajectory.
Then, we project these three representations to the same space by fully-connected layers and add them together to get the final trajectory feature {\small${tr}_i\in\mathbb{R}^{d}$}, as showed in Figure~\ref{fig3}.

Given several trajectory features {\small$\bm{F}_{tr}=\{{tr}_i\}^{N_t}_{i=1}$} in a video, where {\small$N_t$} is unequal for different videos, we employ a trajectory encoder with multi-head self-attention to model the rich trajectory-level interaction,
{\small$\bm{T}=MHSA(\bm{F}_{tr})$},
where {\small$\bm{T}\in\mathbb{R}^{N_t \times d}$} is the refined trajectory feature. 
As illustrated in Figure~\ref{fig3}, our trajectory encoder consists of several multi-head self attention layers~\cite{DBLP:conf/nips/VaswaniSPUJGKP17} and feed-forward layers with skip connection. 

\subsubsection{Language Representations} 
As for language, we use both GloVe features~\cite{DBLP:conf/emnlp/PenningtonSM14} and fine-tuned BERT features~\cite{DBLP:conf/naacl/DevlinCLT19} in different experimental settings. 
A vocabulary set was pre-defined which is composed of top {\small$K$} most frequent words.  
For experiments with GloVe, each word in the set is initialized with word-level pre-trained GloVe representations. 
Following NExT-QA, we also use fine-tuned BERT feature which fine-tunes regular BERT on the dataset by maximizing the correct QA pairs' probability in each multi-choice QA. 
We extract token-wise sentence-level BERT features for each question-answer pair. 
For multi-choice setting, we concatenate the question {\small$Q$} with each candidate answer {\small$A_i$} to form a holistic \emph{query}. 
In order to obtain well contextualized language representation, we apply another GRU to the word embeddings in the query feature {\small$\bm{F}_q$}, 
{\small$\bm{Q},\bm{F}_q^{global}=GRU(\bm{F}_q)$},
where {\small$\bm{Q}\in\mathbb{R}^{L \times d}$} and {\small$\bm{F}_q^{global}\in\mathbb{R}^{d}$} is the global sentence feature from last hidden state.  

\subsection{Feature Alignment and Reasoning}
\label{sec3.2}
\noindent\textbf{Alignment.} For a better alignment among language features, frame-level video features and trajectory features, we propose a cycle-attention module that aligns different features in a circular pattern.
Firstly, we align the trajectory with the language feature, 
\begin{equation}\small
\setlength{\abovedisplayskip}{5pt}
\setlength{\belowdisplayskip}{5pt}
\bm{Q}_{tq} = \underset{q\to t}{Atten}(\bm{Q}, \bm{T}, \bm{T}), \quad
\bm{T}_{tq} = \underset{t\to q}{Atten}(\bm{T}, \bm{Q}, \bm{Q}),
\end{equation}
where ``{\small$\to$}'' means ``attend to''. The {\small$Atten$} operation is introduced in Appendix~\ref{appendix:attention}. Then, we align the frame-level video feature with language feature,
\begin{equation}\small
\setlength{\abovedisplayskip}{5pt}
\setlength{\belowdisplayskip}{5pt}
\bm{Q}_{vq} = \underset{q\to v}{Atten}(\bm{Q}, \bm{V}, \bm{V}), \quad
\bm{V}_{vq} = \underset{v\to q}{Atten}(\bm{V}, \bm{Q}, \bm{Q}).
\end{equation}
We argue that the frame-level video feature and trajectory feature are complementary, since the trajectory feature decouples salient entities from the whole video and the frame-level video feature contains contextual information. 
In addition, there is a natural correspondence relationship between them that trajectory is made up of objects from frames.
Thus, we also align them together,
\begin{equation}\small
\setlength{\abovedisplayskip}{5pt}
\setlength{\belowdisplayskip}{5pt}
\bm{T}_{vt} = \underset{t\to v}{Atten}(\bm{T}, \bm{V}, \bm{V}), \quad
\bm{V}_{vt} = \underset{v\to t}{Atten}(\bm{V}, \bm{T}, \bm{T}).
\end{equation}

\noindent\textbf{Reasoning.} 
After obtaining the aligned features, we conduct heterogeneous graphs, namely TQG, VQG and VTG as shown in Figure~\ref{fig2}, for further reasoning.
Taking the trajectory and question graph TQG as an example, we describe the details of our reasoning module.
The nodes representations {\small$\bm{X}_{tq}$} of TQG are the concatenation of token-wise language embeddings {\small$\bm{Q}_{tq}$} and trajectories features {\small$\bm{T}_{tq}$}.
Thus, each node either represents a word or a trajectory.
We first calculate the value of graph edges represented by an adjacency matrix,
\begin{equation}\small
A_{tq} = softmax(f_{W_p}(\bm{X}_{tq})f_{W_p}(\bm{X}_{tq})^T) + I,
\end{equation}
where {\small$f_{W_p}$} denotes non-linear projection with learnable parameters {\small$W_p$} and {\small$I$} is an identity matrix for skip connection.
Each element of {\small$A_{tq}$} means the correlation between the $i$-th and $j$-th node. 
Then, we apply graph convolution to aggregate and pass message over the nodes. Here we show a single-layer graph convolution operation,
\begin{equation}\small
\setlength{\abovedisplayskip}{5pt}
\setlength{\belowdisplayskip}{5pt}
\bm{X}_{tq}^{(l)} = \sigma(A_{tq}\bm{X}_{tq}^{(l-1)}W^{(l)}),
\end{equation}
where $l$ denotes the $l$-th layer of GCN and $\sigma$ represents an activation function.
To get the final multi-modal representation, we aggregate all the nodes in TQG by weighted pooling with a self-attention, 
\begin{equation}\small
\setlength{\abovedisplayskip}{5pt}
\setlength{\belowdisplayskip}{5pt}
\bm{X}_{tq}^{final} = \sigma(f_{W_t}(\bm{X}_{tq}^{(L)}))\bm{X}_{tq}^{(L)},
\end{equation} 
where {\small$f_{W_t}$} denotes non-linear transformation with learnable parameters {\small$W_t$} and
{\small$\bm{X}_{tq}^{(L)}$} is the output of the last GCN layer.
Similarly, we construct VQG and VTG and conduct graph convolution operations on them to obtain {\small$\bm{X}^{final}_{vq}$} and 
{\small$\bm{X}^{final}_{vt}$}.
\subsection{Answer Prediction}
\label{sec3.3}
Following the multi-choice setting in NExT-QA, we regard the VideoQA task as a multi-modal matching problem, which can easily extend to other multi-modal tasks. Specifically, the candidate answers are concatenated to the corresponding questions and the model scores the concatenated sentences based on the similarities to the video. 

In order to bring in global semantic information, we enhance the three multi-modal features by fusion with the global query feature {\small$\bm{F}_q^{global}$}.
Then, we calculate a score for each candidate question-answer pair using multi-modal compact bilinear fusion (MCB)~\cite{DBLP:conf/emnlp/FukuiPYRDR16},  
\begin{equation}\small
\setlength{\abovedisplayskip}{5pt}
\setlength{\belowdisplayskip}{5pt}
s_* = MCB(\bm{X}_*, \bm{F}_q^{global}),
\end{equation}
where $*$ denotes $tq$, $vq$ and $vt$.
Next, we aggregate scores from different brunches by addition, 
\begin{equation}\small
\setlength{\abovedisplayskip}{5pt}
\setlength{\belowdisplayskip}{5pt}
s = s_{vq}+\lambda_1s_{tq}+\lambda_2s_{vt},
\end{equation}
We adopt a Hinge loss that can maximize the margins between the correct and incorrect QA pairs, 
\begin{equation}\small
\setlength{\abovedisplayskip}{2pt}
\setlength{\belowdisplayskip}{3pt}
L =  \sum_{i=1}^{n-1}{max(0,1+s_i^--s^+)},
\end{equation}
where $n$ is the number of candidate answers and $s^+$ and $s^-$ represent the positive and negative samples.
\subsection{Sample Augmentation}
\label{sec3.4}

\begin{table}[]\small
\centering
\begin{tabular}{lcccc}
\hline
Methods     & Causal & Temp. & Descrip. & Overall \\ \hline
Random      & 20.52  & 20.10  & 19.69  & 20.08   \\
Text Only   & 42.62  & 45.53  & 43.89  & 43.76   \\
Text+Viusal & 42.46  & 46.34  & 45.82  & 44.24   \\
HGA        & 49.53  & 50.74  & 59.33  & 49.74   \\ \hline
Human      & 87.61  & 88.56  & 90.40  & 88.38   \\ \hline
\end{tabular}
\caption{Some VideoQA baselines on NextQA.}
\label{tab1}
\end{table}

\begin{table*}[]\small
\centering
\begin{tabular}{lcccccccccccc}
\hline
\multirow{2}{*}{Methods} &
  \multirow{2}{*}{Text Rep.} &
  \multicolumn{3}{c}{$Acc_C$} &
  \multicolumn{3}{c}{$Acc_T$} &
  \multicolumn{4}{c}{$Acc_D$} &
  \multirow{2}{*}{ACC} \\ \cline{3-12}
      &         & Why   & How         & All   & P\&N & Present & All   & Count       & Loc.   & Other       & All         &       \\ \hline
EVQA  & GloVe   & 28.38 & 29.58       & 28.69 & 29.82      & 33.33   & 31.27 & 43.50       & 43.39       & 38.36       & 41.44       & 31.51 \\
PSAC$^\dagger$ & GloVe   & 35.03 & 29.87       & 33.68 & 30.77      & 35.44   & 32.69 & 38.42       & 71.53       & 38.03       & 50.84       & 36.03 \\
Co-Mem & GloVe   & 36.12 & 32.21       & 35.10 & 34.04      & 41.93   & 37.28 & 39.55       & 67.12       & 40.66       & 50.45       & 38.19 \\
ST-VQA & GloVe   & 37.58 & 32.50       & 36.25 & 33.09      & 40.87   & 36.29 & 45.76       & 71.53       & 44.92       & 55.21       & 39.21 \\
HGA   & GloVe   & 36.38 & 33.82       & 35.71 & 35.83      & 42.08   & 38.40 & {\ul 46.33} & 70.51       & {\ul 46.56} & {\ul 55.60} & 39.67 \\
HME   & GloVe   & 39.14 & 34.70       & 37.97 & 34.35      & 40.57   & 36.91 & 41.81       & {\ul 71.86} & 38.36       & 51.87       & 39.79 \\
HCRN &
  GloVe &
  {\ul 39.86} &
  {\ul 36.90} &
  {\ul 39.09} &
  {\ul 37.30} &
  {\ul 43.89} &
  {\ul 40.01} &
  42.37 &
  62.03 &
  40.66 &
  49.16 &
  {\ul 40.95} \\
\textbf{Ours} &
  GloVe &
  \textbf{43.14} &
  \textbf{39.82} &
  \textbf{42.27} &
  \textbf{40.25} &
  \textbf{47.21} &
  \textbf{43.11} &
  \textbf{46.89} &
  \textbf{74.58} &
  \textbf{52.46} &
  \textbf{59.59} &
  \textbf{45.24} \\ \hline
EVQA  & BERT-FT & 42.31 & 42.90       & 42.46 & 46.68      & 45.85   & 46.34 & 44.07       & 46.44       & 46.23       & 45.82       & 44.24 \\
ST-VQA  & BERT-FT & 45.37 & 43.05       & 44.76 & 44.52      & 51.73   & 49.26 & 43.50       & 65.42       & 53.77       & 55.86       & 47.94 \\
Co-Mem  & BERT-FT & 46.15 & 42.61       & 45.22 & 48.16      & 50.38   & 49.07 & 41.81       & 67.12       & 51.80       & 55.34       & 48.04 \\
HCRN*  & BERT-FT & 46.99 & 42.90       & 45.91 & 48.16      & 50.83   & 49.26 & 40.68       & 65.42       & 49.84       & 53.67       & 48.20 \\
HME    & BERT-FT & 46.52 & {\ul 45.24} & 46.18 & 47.52      & 49.17   & 48.20 & {\ul 45.20} & {\ul 73.56} & 51.15       & 58.30       & 48.72 \\
HGA &
  BERT-FT &
  {\ul 46.99} &
  44.22 &
  {\ul 46.26} &
  {\ul 49.53} &
  {\ul 52.49} &
  {\ul 50.74} &
  44.07 &
  72.54 &
  {\ul 55.41} &
  {\ul 59.33} &
  {\ul 49.74} \\
\textbf{Ours} &
  BERT-FT &
  \textbf{52.81} &
  \textbf{47.44} &
  \textbf{51.40} &
  \textbf{51.11} &
  \textbf{53.70} &
  \textbf{52.17} &
  \textbf{46.89} &
  \textbf{75.25} &
  \textbf{58.03} &
  \textbf{62.03} &
  \textbf{53.30} \\ \hline
\end{tabular}
\caption{Performance (\%) comparisons of state-of-the-art methods on NExT-QA validation set. The best and the second results are bold and underlined respectively. $^\dagger$ means to add motion feature and * means concatenation of question and answer to adapt to BERT representation.}
\label{tab2}
\end{table*}

Although fine-tuned BERT features achieve remarkable results, 
it brings new problem that models answer the question excessively rely on the prior of the question-answer pairs without considering the video content.
It is mainly because the fine-tuning goal is to maximize the probability of the correct QA pair in all the multi-choice QA pair candidates. 
A blind version of VideoQA model was studied by NExT-QA~\cite{DBLP:conf/cvpr/XiaoSYC21} which only considers the question-answer pairs and totally ignores the video inputs.
As shown in Table~\ref{tab1}, the performance of the Text-Only model is surprisingly comparable to the model incorporating the video information. 
We argue that the model devotes to estimating the rationality of question-answer combination or just memorizing the frequency of combinations.
Recent state-of-the-art model HGA~\cite{DBLP:conf/aaai/JiangH20} achieved considerable improvement compared to both Text-Only and Text+Visual models in Table~\ref{tab1}, but there is still a huge gap between the state-of-the-art model and human.
Thus, although the elaborately designed architectures and features have the capacity of reasoning complex interactions, the models always get inferior results. 

Based on this consideration, in order to capitalize on the full potential of the feature and model, we design two effective yet simple sample augmentation ways for better multi-modal alignment and further boost the VideoQA performance.
To be specific, we first increase negative candidate answers (a$^-$)
when computing matching score. This strategy forces the model to
focus more on the minor difference between the correct question-answer pair and others. 
Meanwhile, the model can correspondingly focus on the discriminative video content. 
We then add negative question-answer pairs that are attached to
other videos (qa$^-$). By cooperating with the hinge loss, the video and its affiliated language are drawn closer in feature space and the mismatched pairs are pulled away. In this way, we enhance the multi-modal alignment from a sample perspective. 
In addition, bringing in new negative samples break the models' excessive dependence on language prior, which partially solved the problem caused by the feature. 
In practice, we randomly sample $M$ answers/QA pairs affiliated to other videos in the training set as negative samples.

\section{EXPERIMENTS}
\subsection{Experimental Details}

\noindent\textbf{Dataset.} NExT-QA~\cite{DBLP:conf/cvpr/XiaoSYC21} is a recently designed challenging VideoQA benchmark which advances video question answering from describing to reasoning. 
The dataset contains 5,440 videos where 3870 for training, 570 for validation and 1,000 for testing. 
The videos are selected from the relation dataset VidOR~\cite{shang2019annotating} which contains natural videos of daily life such as outdoor activities and social scenes. Thus they are richer in objects and interactions.
It consists of 47,692 questions where 34,132, 4,996 and 8,564 for training, validation and testing, respectively. 
Almost half of the questions are causal questions which contain questions starting with ``why'' and ``how'', which is a great challenge for VideoQA models to reason about causality.
Temporal questions of inferring temporal actions compose 29\% of the dataset. 
Apart from causal and temporal questions, others are descriptive questions that focus on describing attributes, location and main events in videos.
For multi-choice task that is to select one out of the five candidate answers, NExT-QA sampled four qualified candidates as distracting answers for each question to enhance the hard negatives. 
In a word, NExT-QA goes beyond descriptive QA to benchmark causal and temporal action reasoning in realistic videos and is also rich in object interactions. In addition, several recent state-of-the-art methods are examined on it. 

\noindent\textbf{Evaluation Metric.} We report the accuracy of our model in all   experiments which represents the percentage of correctly answered questions.

\noindent\textbf{Implementation Details.} 
For the training process, we set the number of hidden units $d$ to 256. The batch size is set to 64 and Adam optimizer is used for optimization. The learning rate is set to 0.00005 for GloVe setting and 0.0001 for BERT-FT setting, respectively. For better performance, we reduce the learning rate when a metric has stopped improving. The dropout rate is set to 0.3. We set balance factors $\lambda_1$ and $\lambda_2$ to 0.5 for all the experiments. 

We randomly sample 5 negative samples from the training set for each strategy. We utilize both sample strategies for experiments using GloVe embedding and only use the second strategy for BERT-FT. 
Other details of implementation are given in Appendix~\ref{appendix:implement}.

\begin{table}[]\small
\centering
\begin{tabular}{lccc|c}
\hline
Models & Causal         & Temp.           & Descrip.        & Overall        \\
\hline
ST-VQA & 45.51          & 47.57          & 54.59          & 47.64          \\
Co-Mem & 45.85          & 50.02          & 54.38          & 48.54          \\
HME    & 46.76          & 48.89          & 57.37          & 49.16          \\
L-GCN  & 47.82          & 48.74          & 56.51          & 49.54          \\
HGA    & 48.13          & 49.08          & 57.79          & 50.01          \\
HCRN   & 47.07          & 49.27          & 54.02          & 48.89          \\
HQ-GAU & {\ul 49.04}    & \textbf{52.28} & {\ul 59.43}    & {\ul 51.75}    \\
Ours   & \textbf{50.38} & {\ul 50.88}    & \textbf{61.78} & \textbf{52.41} \\ \hline
\end{tabular}
\caption{Performance(\%) of on NExT-QA test set.}
\label{tab3}
\end{table}

\subsection{Compared Methods}
In Table~\ref{tab2} and Table~\ref{tab3}, we compared our model with other state-of-the-art methods on NExT-QA dataset. 
Among these methods, STVQA~\cite{DBLP:conf/cvpr/JangSYKK17}, PSAC~\cite{DBLP:conf/aaai/LiSGLH0G19} are attention-based methods.
Co-Mem~\cite{DBLP:conf/aaai/GaoZSLLMS19} and HME~\cite{DBLP:conf/cvpr/FanZZW0H19} are memory-based methods. 
L-GCN~\cite{DBLP:conf/aaai/HuangCZDTG20}, HGA~\cite{DBLP:conf/aaai/JiangH20} and HQ-GAU~\cite{DBLP:HQ-GAU} are graph-based methods.

Different from recent elaborately designed complex architectures for VideoQA, we consider the multi-modal alignment from feature and sample perspectives. We simply adopt a heterogeneous graph as the reasoning module and first leverage trajectory feature in VideoQA. We then design two effective yet easy-to-implement sample augmentation methods. Combining both of them, our model achieves the best performance.
 
\noindent\textbf{Results.} The results on NExT-QA validation set and test set are reported in Table~\ref{tab2} and Table~\ref{tab3}, respectively. We can observe that our proposed method achieves new state-of-art performance over all kinds of questions.
In particular, we observe that our method works well even on causal and temporal questions which require more complicated reasoning, \eg,
our method achieves a significant 5.14\% absolute improvement on validation set compared to the second result on causal questions and 1.43\% on temporal questions. 
It should be noticed that HGA also utilizes a heterogeneous graph model for alignment and reasoning which indicates that our trajectory-aware graph model with sample augmentation has the advantage to reason causal and temporal questions over others. 
L-GCN also utilizes a graph network with object feature and our model outperforms it by a large margin on test set as shown in Table~\ref{tab3}.
Recently proposed HQ-GAU also adopt a powerful hierarchical architecture 
with multi-granularity video features that leverages finer object interaction. 
Table~\ref{tab3} shows that our method outperforms HQ-GAU on causal and descriptive questions by 1.34\% and 2.53\%.
For temporal questions, our method gets comparable result with HQ-GAU but is 1.4\% lower than it. It is probably because HQ-GAU has a complicated structure with more parameters and adopts a more effective temporal position embedding.

\begin{table}[t]\small
\centering
\setlength{\tabcolsep}{1.2mm}{
\begin{tabular}{ccc|cccc}
\hline
frame.     & traj.      & aug.       & Causal & Temp. & Discrip. & Overall \\ \hline
           & \ding{51} &            & 46.18  & 48.08 & 57.27    & 48.52   \\
\ding{51} &            &            & 46.49  & 48.76 & 58.94    & 49.16   \\
\ding{51} & \ding{51} &            & 47.18  & 51.18 & 59.33    & 50.36   \\
\ding{51} &            & \ding{51} & 50.44  & 51.30 & 59.85    & 52.18   \\
\ding{51} & \ding{51} & \ding{51} & 51.40  & 52.17 & 62.03    & 53.30   \\ \hline
\end{tabular}
}
\caption{Performance (\%) on validation set in
ablative experiments of trajectory and sample augmentation.}
\label{tab4}
\end{table}
\begin{table}[htbp]\small
\centering
\setlength{\tabcolsep}{1.1mm}{
\begin{tabular}{c|c|cccc}
\hline
Aug.                 & Ablation     & Causal & Temp. & Descrip. & Overall \\ \hline
\multirow{5}{*}{No}  & w/o cyatten. & 46.91  & 48.70 & 59.33    & 49.42   \\
                     & w/o VQG      & 46.18  & 48.08 & 57.27    & 48.52   \\
                     & w/o TQG      & 46.49  & 48.76 & 58.94    & 49.16   \\
                     & w/o VTG      & 46.30  & 50.74 & 57.27    & 49.44   \\
                     & Full         & 47.18  & 51.18 & 59.33    & 50.36   \\ \hline
\multirow{2}{*}{Yes} & traj. GRU    & 50.63  & 53.54 & 60.36    & 53.08   \\
                     & traj. MHSA   & 51.40  & 52.17 & 62.03    & 53.30   \\ \hline
\end{tabular}
}
\caption{Performance (\%) on validation set in
 ablative experiments of model components.}
\label{tab5}
\end{table}

\begin{figure*}
    \centering
    \includegraphics[width=1.0\textwidth]{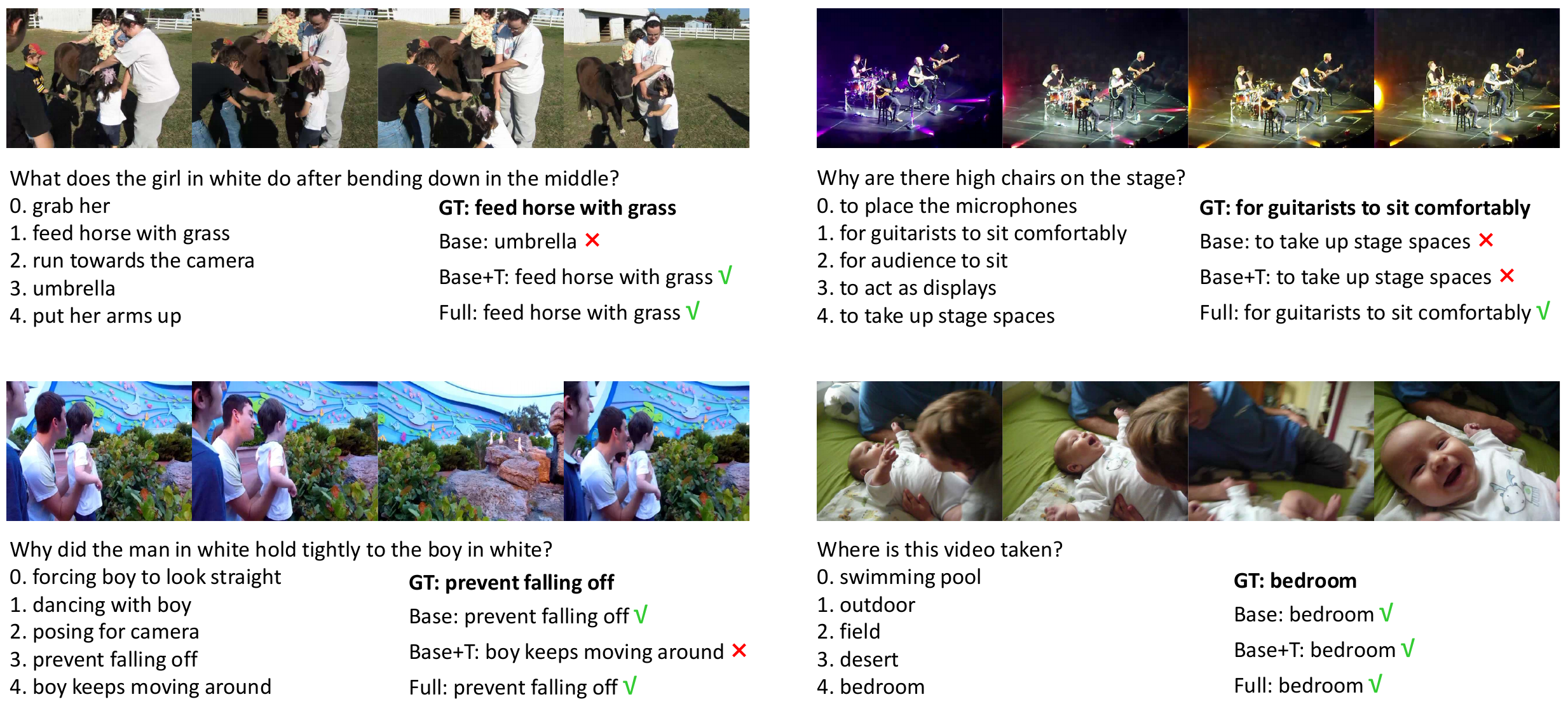}
    \caption{Some qualitative results of our model on NExT-QA validation set. Base: our model without trajectory and sample augmentation. Base+T: Base model with trajectory feature. Full: Base model with trajectory feature and sample augmentation.}
    \label{fig4}
\end{figure*}

\subsection{Ablation Study} 
In this section, we report the results of ablative experiments with different variants to better investigate our approach. 
We first analyze the effect of trajectory feature and sample augmentation method.
Then, we introduce an ablation study conducted on components of our model. 
All the variants in this section are evaluated on NExT-QA validation set.

\noindent\textbf{Ablation on trajectory.}
To exploit the effect of the trajectory, we compared the performance of the models with and without trajectory feature in Table~\ref{tab4}. 
By comparing the second line with the third line, we notice that utilizing trajectory feature improves the accuracy by 1.20\%.  
Comparing the last two lines in Table~\ref{tab4}, the model using trajectory feature outperforms the other by 1.12\% overall accuracy even though the score has already been improved a lot by sample augmentation. 
These results demonstrate the necessity of employing the trajectory feature.
In addition, we give the results of the model only using trajectory feature as visual representation on line 1 in Table~\ref{tab4}. The results indicate that the frame-level feature also plays an important role in multi-modal reasoning. The main reason is that frame-level features provide contextual information which some questions heavily rely on.

\begin{table}[htbp]\small
\centering
\setlength{\tabcolsep}{1mm}{
\begin{tabular}{c|c|cccc}
\hline
Strategy & Text Rep.                & Causal & Temp. & Descrip. & Overall \\ \hline
none       & \multirow{4}{*}{GloVe}   & 36.86  & 37.59 & 54.70    & 39.87   \\
a$^-$        &                          & 41.66  & 43.61 & 57.01    & 44.68   \\
qa$^-$       &                          & 40.20  & 41.07 & 58.43    & 43.31   \\
both     &                          & 42.27  & 43.11 & 59.59    & 45.24   \\ \hline
none       & \multirow{3}{*}{BERT-FT} & 47.18  & 51.18 & 59.33    & 50.36   \\
a$_*^-$       &                          & 47.22  & 49.88 & 59.33    & 49.96   \\
qa$^-$       &                          & 51.40  & 52.17 & 62.03    & 53.30   \\ \hline
\end{tabular}
\caption{Comparisons of different sample augmentation strategies.}
\label{tab6}
}
\end{table}
\noindent\textbf{Ablation on sample augmentation.}
We analyzed the effectiveness of sample augmentation methods in Table~\ref{tab4}.
By comparison of line 2 and line 4 (line 3 vs. line 5), we notice an almost 3\% absolute overall improvement, which indicates that our augmentation methods can boost the performance. 
Considering that it's harder to improve in a higher score range, this indicates the trajectory feature and sample augmentation method could promote each other for better multi-modal alignment. 

We also explore the different augmentation strategies in Table~\ref{tab6}. 
The a$^-$ represents that we sample negative answers from other questions and concatenate them with the original question. The qa$^-$ means that we sample negative question-answer pairs attached to other videos.
For experiments with GloVe embedding as text representation, we can find that each strategy improved the accuracy by a large margin and using both strategies can further boost the performance.
With regards to BERT-FT as text representation, we cannot directly apply strategy a$^-$ because BERT features are sentence-level holistic feature for question-answer pairs where the question parts vary across different pairs.
So we averaged the question features as a unified question representation and concatenated randomly sampled answers (a$_*^-$ in Table~\ref{tab6}).
However, we can observe that the accuracy barely changed.
This may be because the average operation harms the integrity of sentence representation especially the sentence matching information that [CLS] embedding contains.
Thus we only used the second strategy qa$^-$ for BERT-FT and the performance is improved a lot even so.
We also analyzed the effect of the number of negative samples. The performance grows when the number of negative samples increases. When the number is more than 15, the performance would barely change.

\noindent\textbf{Ablation on trajectory encoder.}
On the bottom section of table~\ref{tab5}, we studied the effect of trajectory encoder MHSA.
By replacing our MHSA with a GRU with temporal and semantic embedding, the performance drops by 0.77\% on causal questions and 1.67\% on descriptive questions making a overall 0.22\% decline, which demonstrates the global interactions modeling ability of MHSA.
For temporal questions, there is a 1.37\% improvement which indicates that the RNN architecture is better to capture sequential information.

\noindent\textbf{Ablation on model components.}
We analyzed the model components on the top part of Table~\ref{tab5}.
By removing each graph and cycle-attention, performance of the model all dropped. 
The results demonstrate that all parts of the architecture play an important role in alignment and reasoning.

\subsection{Qualitative Analysis}
\label{qualitative}
We show some qualitative results on NExT-QA validation set in Figure~\ref{fig4}. The results of three models with different configurations are visualized, \ie, Base: the baseline without trajectory feature and sample augmentation, Base+T: add trajectory feature to Base, and Full: our full model with both trajectory and sample augmentation. We notice that Base+T and Full model perform better than Base in most cases, which demonstrates that both feature and strategies are helpful. In the bottom-left case, we surprisingly found that Base+T predicted a wrong answer whereas Base answered correctly. However, the candidate answer 4 ``boy keeps moving around'' seems hardly to be a wrong answer to the question.

\section{CONCLUSION}
In this paper, we explored multi-modal alignment in VideoQA from feature and sample perspectives.
From the view of feature, we first leverage video trajectory features in VideoQA to bridge the semantic gap between the sub-components of the video and the language.
Moreover, in order to better utilize the trajectory feature, we propose a graph-based model which is capable of alignment and reasoning over heterogeneous representations.
From the view of sample, we propose two sample augmentation strategies to further enhance the cross-modal correspondence ability of our model.
The promising results on challenging NExT-QA dataset have exhibited the causal and temporal reasoning ability of our method. In the future, we will further explore a better way to take advantage of trajectory information considering its significant potential.

\section*{Limitations}
Although video trajectories are effective on VideoQA and other video understanding tasks, the model is sensitive to 
the quality of trajectories. 
The object detection and tracking methods are of vital important to the quality of trajectories. 
Using a weak tracking method may introduce noises which can be harmful to the performance. 
The training augmentation strategies are naturally suitable for multi-choice setting, however for open-ended setting, further work needs to be done to adapt. 


\section*{Acknowledgements}
This work was supported by the National Key Research \& Development Project of China (2021ZD0110700), the National Natural Science Foundation of China (U19B2043, 61976185), Zhejiang Natural Science Foundation (LR19F020002), Zhejiang Innovation Foundation (2019R52002), and the Fundamental Research Funds for the Central Universities (226-2022-00087).

\bibliography{anthology,custom}
\bibliographystyle{acl_natbib}

\appendix

\section*{Appendix}

\section{Attention Operation}
\label{appendix:attention}
Attention can be generalised to compute a weighted sum of the values dependent on the query and the corresponding keys. Since the query determines which values to focus on, we can say that the query attends to the values.
Given a query $\tilde{Q}$ and a set of key-value pairs ($\tilde{K}$, 
$\tilde{V}$), dot-product attention adopted by Transformer~\cite{DBLP:conf/nips/VaswaniSPUJGKP17} computes the alignment weights using dot-production of $\tilde{Q}$ and $\tilde{K}$ as shown in the right part of Figure~\ref{fig3},
\begin{equation}\small
Atten(\tilde{Q}, \tilde{K}, \tilde{V}) =  softmax(\frac{(\tilde{Q}W_h^{\tilde{Q}})(\tilde{K}W_h^{\tilde{K}})^T}{\sqrt{d_h}}) \tilde{V}W_h^{\tilde{V}}, 
\end{equation}
where $W_h^{\tilde{Q}}$, $W_h^{\tilde{K}}$ and $W_h^{\tilde{V}}$ are trainable projection matrices and $\sqrt{d_h}$ is a scaling factor that
prevents softmax function from excessively large with keys of higher dimensions.

\section{Implementation Details}
\label{appendix:implement}

\textbf{Frame-level feature details.}
We uniformly split each video into 16 segments and each segment has 16 consecutive frames. 
We utilize a ResNet-101~\cite{DBLP:conf/cvpr/HeZRS16} pre-trained on ImageNet~\cite{DBLP:conf/cvpr/DengDSLL009} to extract per-frame appearance feature of 2048-D. As for the 2048-D motion feature, we utilize an I3D ResNeXt-101~\cite{DBLP:conf/cvpr/HaraKS18} pre-trained on Kinetic~\cite{DBLP:journals/corr/KayCSZHVVGBNSZ17} as mainstream framework. 

\noindent\textbf{Trajectory-level feature details.} We sample video at a rate of 1fps.
We adopt Faster-RCNN~\cite{DBLP:conf/nips/RenHGS15}  trained on open-Images as the object detector, which uses Inception Resnet V2 as the image feature extractor, containing 600 classes. 
We use a dynamic programming algorithm improved from sequence NMS to associate bounding boxes that belong to the same object and generate trajectories.
This tracking method consists of two steps: graph building and trajectory selection and we refer readers to~\cite{DBLP:conf/mm/XieR020} for more details.

\noindent\textbf{Language representation details.} We first extract tokens from sentences. Then we employ the GloVe~\cite{DBLP:conf/emnlp/PenningtonSM14} pre-trained on Wikipedia to obtain 300-D embedding for each word token. The maximum length of question-answer pairs is set to 37. We truncated the sentences longer than the max length and padded the shorter ones with zeros.
For the BERT-FT setting, we directly utilized finetuned BERT feature provided by NExT-QA~\cite{DBLP:conf/cvpr/XiaoSYC21}.
Each answer is appended to the question as a global sentence. A BERT build-in tokenizer is used to obtain the tokenized representation of the sentence. 
Then the tokens are organized by the format: [CLS] question [SEP] candidate answer [SEP].

\end{document}